\documentclass{article}

\usepackage{microtype}
\usepackage{graphicx}
\usepackage{subfigure}
\usepackage{booktabs} 
\usepackage{amsmath}
\usepackage{bbm}

\usepackage{nicefrac}  
\usepackage{lipsum}
\usepackage{graphicx}
\usepackage{amsfonts}
\usepackage{mathabx}
\usepackage{wrapfig}

\usepackage{xcolor} 
\usepackage[accepted]{icml2020}

\pdfoutput=1

\usepackage[ruled,vlined,algo2e]{algorithm2e}
\usepackage{hyperref}   
\usepackage{graphicx}
\usepackage{lipsum}  
\usepackage{wrapfig}
\usepackage{xcolor}
\usepackage{float}

\newcommand{\rs}{\mathbf{s}}
\newcommand{\ra}{\mathbf{a}}
\newcommand{\rst}{\mathbf{s}_{1:T}}
\newcommand{\rat}{\mathbf{a}_{1:T}}
\newcommand{\rsh}{\mathbf{s}_{t:T}}
\newcommand{\rah}{\mathbf{a}_{t:T}}
\newcommand{\optimal}{\mathcal{O}}
\newcommand{\optimalt}{\mathcal{O}_{1:T}}

\icmltitlerunning{Control as Hybrid Inference}

\begin{document}

\twocolumn[
\icmltitle{Control as Hybrid Inference}

\begin{icmlauthorlist}
\icmlauthor{Alexander Tschantz}{sackler,sussex}
\icmlauthor{Beren Millidge}{edinburgh}
\icmlauthor{Anil K. Seth}{sussex,sackler,cifar}
\icmlauthor{Christopher L. Buckley}{sussex}
\end{icmlauthorlist}

\icmlaffiliation{edinburgh}{School of Informatics, University of Edinburgh, United Kingdom}
\icmlaffiliation{sussex}{Evolutionary and Adaptive Systems Group, University of Sussex, United Kingdom}
\icmlaffiliation{sackler}{Sackler Centre for Conscsiousness Science}
\icmlaffiliation{cifar}{Canadian Institute for Advanced Research}
\icmlcorrespondingauthor{Alexander Tschantz}{tschantz.alec@gmail.com}
\icmlkeywords{Reinforcement Learning, Control as Inference, Variational Inference}

\vskip 0.3in
]

\printAffiliationsAndNotice  

\begin{abstract}
  The field of reinforcement learning can be split into model-based and model-free methods. 
  Here, we unify these approaches by casting model-free policy optimisation as amortised variational inference, and model-based planning as iterative variational inference, within a `control as hybrid inference' (CHI) framework. 
  We present an implementation of CHI which naturally mediates the balance between iterative and amortised inference.
  Using a didactic experiment, we demonstrate that the proposed algorithm operates in a model-based manner at the onset of learning, before converging to a model-free algorithm once sufficient data have been collected. 
  We verify the scalability of our algorithm on a continuous control benchmark, demonstrating that it outperforms strong model-free and model-based baselines.
  CHI thus provides a principled framework for harnessing the sample efficiency of model-based planning while retaining the asymptotic performance of model-free policy optimisation.
\end{abstract}

\vspace{-0.6cm}
\section{Introduction}
\label{sec:introduction}

Reinforcement learning (RL) algorithms can generally be divided into model-based and model-free approaches.
Model-based algorithms utilise a model of the environment to facilitate action selection, allowing them to generalize knowledge to new tasks and learn from a handful of trials \cite{chua2018deep}.
In contrast, model-free algorithms learn a policy (or value function) directly from experience and generally show increased asymptotic performance relative to their model-based counterparts \cite{mnih2015human}.
However, such algorithms tend to be substantially less sample efficient.
It would therefore be helpful to have a principled method for combining these approaches, harnessing the sample efficiency of model-based RL and the asymptotic performance of model-free RL \cite{wang2019exploring}.

In this work, we show that the \emph{control as inference} framework \cite{levine2018reinforcement,rawlik2013stochastic,ziebart2010modeling,abdolmaleki2018maximum} provides a principled methodology for combining model-based and model-free RL. 
This framework casts decision making as probabilistic inference, enabling researchers to derive principled (Bayesian) objectives and draw upon a wide array of approximate inference techniques.
While the framework encompasses many different methods, they all share the goal of inferring a posterior distribution over actions, given a probabilistic model that is conditioned on observing `optimal' trajectories.
  
Since computing the posterior distribution over actions is generally analytically intractable, variational methods are often used to implement approximate inference \cite{beal2003variational}.
Here, we highlight a distinction between \emph{amortised} and \emph{iterative} approaches to variational inference \cite{kim2018semi}, and show that, in the context of control as inference, amortised inference naturally corresponds to model-free policy optimisation, whereas iterative inference naturally corresponds to model-based planning. 

Leveraging these insights, we propose \emph{control as hybrid inference} (CHI), a framework for combining amortised and iterative inference in the context of control. 
This framework proposes two inference algorithms -- one amortised and one iterative -- which work collaboratively to recover an (approximate) posterior over sequences of actions.
To combine these processes, we implement an algorithm in which amortised inference sets the initial conditions for a subsequent phase of iterative inference, leading to a natural and adaptive interaction between the two inference approaches.

We utilise a didactic experiment to investigate the interaction between amortised and iterative inference over the course of learning and explore how this is affected when environmental contingencies change. 
We find that iterative inference dominates action selection when amortised predictions are uncertain, such as at the onset of learning, and that amortised inference determines action selection when sufficient data have been collected.
We demonstrate the scalability of our algorithm using a high-dimensional control benchmark and demonstrate that it outperforms strong model-based and model-free baselines, both in terms of sample efficiency and asymptotic performance.
These results suggest that CHI could provide a principled framework for combining the sample efficiency of model-based planning with the asymptotic performance of model-free policy optimisation.

\vspace{-0.25cm}
\section{Background}
\label{sec:background}

We consider a finite-horizon Markov decision process (MDP) defined by a tuple $\{\mathcal{S}, \mathcal{A}, p_{\texttt{env}}, r\}$, where $\rs \in \mathcal{S}$ denotes states, $\ra \in \mathcal{A}$ denotes actions, $p_{\texttt{env}}(\rs_{t+1}|\rs_{t}, \ra_t)$ is the environment's dynamics and $r(\rs_t, \ra_t)$ is the reward function. 
Traditionally, RL problems look to identify the policy $p_{\theta}(\ra_t |\rs_t)$ which maximizes the expected sum of reward $\mathbb{E}_{p_{\theta}(\tau)}\big[\sum_{t=1}^T r(\rs_t, \ra_t)\big]$, where $\theta$ are the policies parameters, $\tau$ denotes a trajectory $\tau = \{(\rs_t, \ra_t)\}_{t=1}^T$, and $p_{\theta}(\tau)$ denotes the probability of trajectories under a policy, $p_{\theta}(\tau) = p(\rs_1) \prod_{t=1}^T p_{\theta}(\ra_t| \rs_t)p_{\texttt{env}}(\rs_{t+1}|\rs_{t}, \ra_t)$. 

\vspace{-0.2cm}
\paragraph{Control as Inference} To reformulate the problem of RL in the language of probability theory, we introduce an auxillary `optimality' variable $\optimal \in [0, 1]$, where $\optimal_t = 1$ denotes that time step $t$ was optimal (we drop $= 1$ for conciseness).
We assume agents encode a generative model over trajectories and optimality variables:
\vspace{-0.25cm}
\begin{equation}
  p(\tau, \optimalt ) = p(\rs_1) \prod_{t=1}^T p(\optimal_t|\rs_t, \ra_t) p_{\lambda}(\rs_{t+1}| \rs_t, \ra_{t}) p(\ra_t)
\end{equation}
\vspace{-0.05cm}
where $\lambda$ are parameters of the dynamics model, which may be learned in a model-based context. 
We assume an uninformative action prior $p(\ra_t) = \frac{1}{|\mathcal{A}|}$.
The optimality likelihood $p(\optimal_t|\rs_t, \ra_t)$ describes the probability that some state-action pair $(\rs_t, \ra_t)$ is optimal and is defined as $p(\optimal_t|\rs_t, \ra_t) = \mathrm{exp}\big(r(\rs_t, \ra_t)\big)$ \cite{levine2018reinforcement}.

The goal of control as inference is to maximise the marginal-likelihood of optimality $p(\optimalt)$.
While it is generally intractable to evaluate this quantity directly, it is possible to construct a variational lower bound $\mathcal{L}$ which can be evaluated and optimised through variational inference \cite{jordan1999introduction}.
To achieve this, we introduce an arbitrary distribution over trajectories $q(\tau) = q(\rs_1) \prod_{t=1}^T q(\rs_{t+1}| \rs_t, \ra_{t}) q_{\theta}(\ra_t|\rs_t) $, which we refer to as an \emph{approximate posterior}.
The variational lower bound $\mathcal{L}$ is then given by (see Appendix \ref{ap:bound} for a derivation):
\begin{equation}
\label{eq:bound}
    \mathcal{L} = - D_\mathrm{KL}\Big(q(\tau) \ \Vert \ p(\tau| \optimalt)\Big) \leq \log  p(\optimalt) 
\end{equation}
Maximising $\mathcal{L}$ with respect to the parameters of the approximate posterior provides a tractable method for maximising the (log) marginal-likelihood of optimality. 
We can further simplify Eq. \ref{eq:bound} by fixing $q(\rs_1) = p(\rs_1)$ and $q(\rs_{t+1}| \rs_t, \ra_{t}) = p_{\lambda}(\rs_{t+1}| \rs_t, \ra_{t})$, giving (see Appendix \ref{ap:simplifcation} for a derivation):
\vspace{-0.3cm}
\begin{equation}
  \label{eq:simple-bound}
  \begin{aligned}
      \mathcal{L} = \mathbb{E}_{q(\tau)}\Big[\sum_{t=1}^T r(\rs_t, \ra_t)\Big] + \mathbf{H}\Big[q_{\theta}(\rat|\rst)\Big]
  \end{aligned}
\end{equation}
where $\mathbf{H}[\cdot]$ is the Shannon entropy. 
The inclusion of the action entropy term provides several benefits, including a mechanism for offline learning \cite{nachum2017bridging}, improved exploration and increased algorithmic stability.
Empirically, algorithms derived from the control as inference framework often outperform their non-stochastic counterparts \cite{haarnoja2018soft}.

\vspace{-0.3cm}
\paragraph{Iterative Inference} Equation \ref{eq:bound} demonstrates that control as inference corresponds to a process of variational inference. 
In the wider literature on variational inference, a key distinction is made between \emph{iterative} and \emph{amortised} approaches. 
Iterative approaches to variational inference directly optimise the parameters of the approximate posterior $\theta$ in order to maximise $\mathcal{L}$, a process which is carried out iteratively for each data point.
Examples of this approach include stochastic variational inference \cite{hoffman2013stochastic}, variational message passing \cite{winn2005variational}, belief propagations \cite{weiss2000correctness} and variational expectation-maximisation \cite{marino2018iterative}.
Within the control as inference framework, the approximate posterior is over actions, and when considering an approximate posterior over \textit{sequences} of actions, several model-based planning algorithms can be cast as a process of iterative inference \cite{okada2019variational, piche2018probabilistic, williams2017information, friston2015active, tschantz2019scaling}.

\vspace{-0.3cm}
\paragraph{Amortised Inference} In contrast, amortised variational inference learns a parameterised function $f_{\phi}(\cdot)$ which maps directly from data $\mathbf{x}$ to the parameters of the approximate posterior $\theta \leftarrow f_{\phi}(\mathbf{x})$.
Amortised inference models are learned by optimising the parameters $\phi$ in order to maximise $\mathcal{L}$, an optimisation that takes place over the available dataset $\mathcal{D}$. Amortised variational inference form the basis of popular methods such as variational autoencoders \cite{kingma2013auto}.
In the context of control as inference, the parameterised function corresponds to a \emph{policy}, and the approximate posterior is again over actions.
This approach is closely related the field of maximum-entropy RL \cite{eysenbach2019if}, which has inspired several influential model-free algorithms \cite{levine2018reinforcement, haarnoja2018soft, abdolmaleki2018maximum}. 

\vspace{-0.3cm}
\section{Control as Hybrid Inference}
\label{sec:chi}
In this section, we introduce the \emph{control as hybrid inference} (CHI) framework.
Like the control as inference framework, CHI proposes that agents infer an approximate posterior over actions, given a generative model that is conditioned on `optimality'.
However, CHI additionally proposes that inference is achieved via two processes -- an amortised process which maps from states to the parameters of an approximate posterior over actions, and an iterative process which directly updates the parameters of the approximate posterior in an iterative manner.
To ensure consistency between these processes, both amortised and iterative inference utilise the same generative model and optimise the same variational objective. 
By utilising the correspondence between \textbf{(i)} amortised inference and policy optimisation, and \textbf{(ii)} iterative inference and planning, we demonstrate that this perspective allows for a principled combination of model-based and model-free RL. 

\vspace{-0.3cm}
\paragraph{Iterative Inference Algorithm} We consider an iterative inference algorithm which optimises an approximate posterior over action \emph{sequences} of a fixed horizon $H$ extending from the current time step $t$, $q(\rah; \theta)$, where we have used $T = t + H$ to simplify notation.\footnote{For clarity, we adopt notation $q(\ra; \theta)$ to denote an iterative posterior and $q_{\phi}(\ra|\rs; \theta)$ to denote an amortised inference model.}
We consider this distribution to be a time-dependent diagonal Gaussian, $q(\rah; \theta) = \mathcal{N}(\rah; \mu_{t:T}, \mathrm{diag} \ \sigma_{t:T}^2)$, where $\theta = \{ \mu_{t:T}, \sigma_{t:T}^2 \}$.

At each time step $t$, agent's observe the state of the environment $\rs_t$. 
Iterative inference then proceeds by iteratively updating the parameters of $q(\rah; \theta)$ in order to maximise $\mathcal{L}$. 
As demonstrated in \cite{okada2019variational}, this can be achieved by utilising  of mirror descent \cite{bubeck2014convex, okada2018acceleration}, providing the following update rule:
\begin{equation}
\begin{aligned}
  \label{eq:vimpc}
  q^{(i+1)}(\rah; \theta) \leftarrow \frac{ q^{(i)}(\rah; \theta) \cdot \mathcal{W}(\rah)\cdot q^{(i)}(\rah; \theta)}{\mathbb{E}_{q^{(i)}(\rah; \theta)}\Big[\mathcal{W}(\rah) \cdot q^{(i)}(\rah; \theta)\Big]} 
\end{aligned}
\end{equation}
where $i$ denotes the current iteration and $\mathcal{W}\big(\rah\big) = \mathbb{E}_{q(\rsh|\rah, \rs_t)}\big[r(\tau)\big]$. 
After $I$ iterations, the mean of the approximate posterior over action $\mu_{t:T}$ is returned and the first action $\mu_t$ from this sequence is executed. 
Equation \ref{eq:vimpc} is a Bayesian generalisation of model predictive path integral control (MPPI) \cite{williams2017information}, a popular method for model predictive control (MPC) \cite{camacho2013model}.

\vspace{-0.3cm}
\paragraph{Amortised Inference Algorithm} Our amortised inference algorithm infers an approximate posterior over the \emph{current} action $q_{\phi}(\ra_t|\rs_t; \theta)$. This implies that $q_{\phi}(\ra_t|\rs_t; \theta) = \mathcal{N}(\ra_t; \mu_t, \mathrm{diag} \ \sigma_t^2)$, where $\theta = \{\mu_t, \sigma_t^2\}$.
Rather than optimising $\theta$ directly, amortised inference employs a parameterised function $f_{\phi}(\rs_t)$ which maps directly from $\rs_t$ to $\theta$.
The parameters of this function $\phi$ are then updated in order to maximise the variational bound $\mathcal{L}$.
This optimisation takes place in a batched fashion over the available dataset $\mathcal{D} = \{(\rs_t, \ra_t, r(\rs_t, \ra_t), \rs_{t+1} )\}_{t=1}^B$, where $B$ is the size of the dataset, such that the optimisation problem $\mathrm{argmax}_{\phi} \mathcal{L}(\phi)$ is augmented to $\mathrm{argmax}_{\phi} \mathbb{E}_{\mathcal{D}}\big[\mathcal{L}(\phi)\big]$.

In the current work, we utilise the Soft Actor-Critic (SAC) algorithm \cite{haarnoja2018soft} to optimise $\phi$. 
Rather than directly differentiating the variational bound $\mathcal{L}$, SAC employs a message passing approach.
We refer readers to \cite{haarnoja2018soft} for a description of the SAC algorithm, and \cite{levine2018reinforcement} for a description of its relationship to variational inference.

\begin{figure*}[t!]
  \centering
  \includegraphics[width=\textwidth]{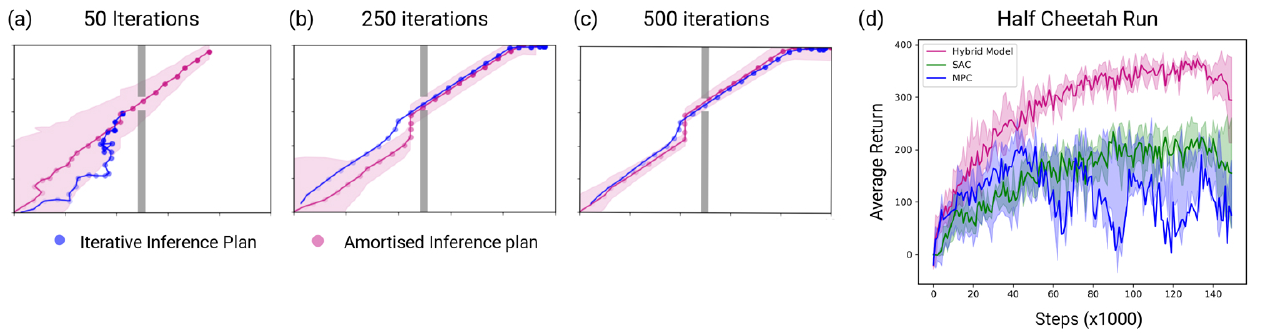}
  \caption{\textbf{(a - c)}: Amortised predictions of $q_{\phi}(\rah|\rsh; \theta)$ are shown in red, where $\bullet$ denote the expected states, shaded areas denote the predicted actions variance at each step, and the expected trajectory recovered by iterative inference is shown in blue.
  At the onset of learning (\textbf{a}), the amortised predictions are highly uncertain, and thus have little influence on the final approximate posterior. As the amortised model $f_{\phi}(\cdot)$ learns (\textbf{b}), the certainty of the amortised predictions increase, such that the final posterior remains closer to the initial amortised guess. At convergence, (\textbf{c}), the iterative phase of inference has negligible influence on the final distribution, suggesting convergence to a model-free algorithm. \textbf{(d)} Here, we compare our algorithm to its constituent components -- the soft-actor critic (SAC) and an MPC algorithm based on the cross-entropy method (CEM). These results demonstrate that the hybrid model significantly outperforms both of these methods.}
  \label{fig:two}
\end{figure*}

\vspace{-0.3cm}
\paragraph{Combining Amortised \& Iterative Inference} We now consider how the amortised and iterative processes can be combined into a single inference algorithm.
In our implementation, amortised inference provides an `initial guess' at the approximate posterior which is then refined by a subsequent phase of iterative inference.
Formally, at each time step $t$, the parameters $\theta$ are initialised by the amortised mapping $\theta = f_{\phi}(\rs_t)$, and then iteratively updated according to Eq. \ref{eq:vimpc}. 

An immediate challenge for this approach is that amortised inference considers an approximate posterior over a \emph{single} action $q_{\phi}(\ra_t|\rs_t; \theta)$, whereas iterative inference considers an approximate posterior over a \emph{sequence} of actions $q(\rah; \theta)$.
To address this challenge, we adapt the amortised algorithm to predict a sequence of actions $q_{\phi}(\rah|\rsh; \theta)$ by
applying the factorisation $q_{\phi}(\rah|\rsh; \theta) = \prod_{t'=t}^T q_{\phi}(\ra_{t'}|\rs_{t'};\theta)$.\footnote{An alternative approach would be to amortise the action sequence directly, such that $f_{\phi}(\rs_t)$ predicts the parameters over a sequence of actions $\theta = \{ \mu_{t:T}, \sigma_{t:T}^2 \}$.}
Thus, $f_{\phi}(\cdot)$ still predicts the parameters of a distribution over current actions $\theta = \{ \mu_{t}, \sigma_{t}^2 \}$.
However, this factorisation raises a separate  issue, in that it requires knowledge of $\rsh$, which are future states and thus unknown.
To overcome this issue, we utilise the learned transition model $p_{\lambda}(\rs_{t+1}|\rs_t, \ra_t)$ (described in Appendix \ref{ap:model-details}) to predict the trajectory of future states $\rsh$. 
Let $p_{\phi}(\tau)$ denote the probability of trajectories under the amortised policy:\footnote{Note that this is not equivalent to $q(\tau)$, which defines the probability of trajectories under the CHI algorithm.}
\vspace{-0.4cm}
\begin{equation}
\begin{aligned}
\label{eq:policy-dist}
  p_{\phi}(\tau) &= p(\rs_t) \prod_{t'=t}^T p_{\lambda}(\rs_{t'+1}|\rs_{t'}, \ra_{t'}) q_{\phi}(\ra_{t'}|\rs_{t'}; \theta) \\
\end{aligned}
\end{equation}
where we have assumed $p(\rs_t) = \delta(\rs_t)$.
We can then recover the desired distribution over actions $q_{\phi}(\rah|\rsh; \theta)$ with parameters $\theta = \{ (\mu_{t'}, \sigma_{t'}^2) \}^T_{t'=t}$, which can then be used to specify the parameters of a time-dependent diagonal Gaussian $\theta = \{ \mu_{t:T}, \sigma_{t:T}^2 \}$. 
These parameters are then used as the initial distribution for the iterative phase of inference.  An overview of the proposed method is provided in Algorithm \ref{ap:algo}, and discuss our solution to the \emph{data bias} issue in Appendix \ref{ap:data-bias}

\vspace{-0.3cm}
\section{Experiments}
\label{sec:experiments}
\paragraph{Didatic experiment} To demonstrate the characteristic dynamics of our algorithm, we utilise a simple 2D point mass environment in which an agent must navigate to a goal (top right-hand corner), with the additional complexity of traversing through a small hole in a wall (see Appendix \ref{ap:experimental-details} for details). We compare the amortised predictions of $q(\rah|\rsh)$ to the final posterior recovered by iterative inference over the course of learning. These results demonstrate that when the amortised predictions are uncertain, such as at the start of learning, the posterior inferred by iterative inference is relatively unaffected by the amortised predictions, suggesting the model acts in a primarily model-based manner. Once sufficient data has been collected and the amortised predictions are precise, the iterative phase of inference has a negligible effect on the final distribution, suggesting a gradual convergence from a model-based to a model-free algorithm.
\vspace{-0.3cm}
\paragraph{Continious control} As a proof of principle, we demonstrate our algorithm can scale to complex tasks by evaluating performance on the challenging Half-Cheetah task (see Appendix \ref{ap:experimental-details} for details). We compare the CHI algorithm to the model-free SAC and a model-based planning algorithm which utilises the cross-entropy method for trajectory optimisation. These results demonstrate that CHI outperforms both baselines in terms of sample efficiency and asymptotic performance. Note that the performance of MPC is lower than what has been reported in previous literature. We believe this is due to the fact that we utilised fewer parameters relative to prior work. These results suggest that a hybrid approach can help stabilize planning algorithms, enabling comparable performance with reduced computational overhead. Indeed, there is no difference between the MPC algorithm and the iterative component of the CHI algorithm, thus establishing the benefit of a hybrid approach.
\vspace{-0.45cm}
\section{Conclusion}
\label{sec:conclusion}

In this work, we have introduced \emph{control as hybrid inference} (CHI), a framework for combining model-free policy optimisation and model-based planning in a probabilistic setting, and provided proof-of-principle demonstrations that CHI retains the sample efficiency of model-based RL and the asymptotic performance of model-free RL. We finish by highlighting several additional benefits afforded by the CHI framework. 
First, initialising a model-based planning algorithm with an `initial guess' significantly reduces the search space.
Moreover, by employing amortised inference schemes that utilise a value function, it should be possible to estimate the value of actions beyond the planning horizon. 
Furthermore, the fact that the certainty of amortised predictions increases over the course of learning suggests the possibility of terminating iterative inference once a suitable threshold (in terms of the standard deviation) has been reached, which would decrease the computational cost of model-based planning.
We also expect that the relative influence of the two algorithms will be adaptively modulated in the face of changing environmental contingencies, as confirmed in preliminary experiments. 
Finally, the CHI framework provides a formal model of the hypothesis that model-free and model-based mechanisms coexist and compete in the brain according to their relative uncertainty \cite{niv2006normative, daw2005uncertainty}, as well as explaining \emph{habitization}, or the gradual transition from goal-directed to habitual action after sufficient experience. \cite{glascher2010states}.

While we have proposed one implementation of CHI based on initialisation, several alternatives exist. For instance, the amortised component could be incorporated as an action prior in the graphical model. Moreover, while we have implemented CHI using particular algorithms, these could be replaced by a wide range of state-of-the-art RL algorithms. This is possible due to the observation that, under a control as inference perspective, model-based planning and model-free policy optimisation generally correspond to iterative and amortised inference, respectively (Millidge et al., in press).
\clearpage


\bibliography{refs}

\begin{thebibliography}{50}
\providecommand{\natexlab}[1]{#1}
\providecommand{\url}[1]{\texttt{#1}}
\expandafter\ifx\csname urlstyle\endcsname\relax
  \providecommand{\doi}[1]{doi: #1}\else
  \providecommand{\doi}{doi: \begingroup \urlstyle{rm}\Url}\fi

\bibitem[Abdolmaleki et~al.(2018)Abdolmaleki, Springenberg, Tassa, Munos,
  Heess, and Riedmiller]{abdolmaleki2018maximum}
Abdolmaleki, A., Springenberg, J.~T., Tassa, Y., Munos, R., Heess, N., and
  Riedmiller, M.
\newblock Maximum a posteriori policy optimisation.
\newblock \emph{arXiv preprint arXiv:1806.06920}, 2018.

\bibitem[Beal et~al.(2003)]{beal2003variational}
Beal, M.~J. et~al.
\newblock \emph{Variational algorithms for approximate Bayesian inference}.
\newblock university of London London, 2003.

\bibitem[Bubeck(2014)]{bubeck2014convex}
Bubeck, S.
\newblock Convex optimization: Algorithms and complexity.
\newblock \emph{arXiv preprint arXiv:1405.4980}, 2014.

\bibitem[Camacho \& Alba(2013)Camacho and Alba]{camacho2013model}
Camacho, E.~F. and Alba, C.~B.
\newblock \emph{Model predictive control}.
\newblock Springer Science \& Business Media, 2013.

\bibitem[Che et~al.(2018)Che, Lu, Tucker, Bhupatiraju, Gu, Levine, and
  Bengio]{che2018combining}
Che, T., Lu, Y., Tucker, G., Bhupatiraju, S., Gu, S., Levine, S., and Bengio,
  Y.
\newblock Combining model-based and model-free rl via multi-step control
  variates.
\newblock 2018.

\bibitem[Chebotar et~al.(2017)Chebotar, Hausman, Zhang, Sukhatme, Schaal, and
  Levine]{chebotar2017combining}
Chebotar, Y., Hausman, K., Zhang, M., Sukhatme, G., Schaal, S., and Levine, S.
\newblock Combining model-based and model-free updates for trajectory-centric
  reinforcement learning.
\newblock In \emph{Proceedings of the 34th International Conference on Machine
  Learning-Volume 70}, pp.\  703--711. JMLR. org, 2017.

\bibitem[Chua et~al.(2018)Chua, Calandra, McAllister, and Levine]{chua2018deep}
Chua, K., Calandra, R., McAllister, R., and Levine, S.
\newblock Deep reinforcement learning in a handful of trials using
  probabilistic dynamics models.
\newblock In \emph{Advances in Neural Information Processing Systems}, pp.\
  4754--4765, 2018.

\bibitem[Cremer et~al.(2018)Cremer, Li, and Duvenaud]{cremer2018inference}
Cremer, C., Li, X., and Duvenaud, D.
\newblock Inference suboptimality in variational autoencoders.
\newblock \emph{arXiv preprint arXiv:1801.03558}, 2018.

\bibitem[Daw et~al.(2005)Daw, Niv, and Dayan]{daw2005uncertainty}
Daw, N.~D., Niv, Y., and Dayan, P.
\newblock Uncertainty-based competition between prefrontal and dorsolateral
  striatal systems for behavioral control.
\newblock \emph{Nature neuroscience}, 8\penalty0 (12):\penalty0 1704--1711,
  2005.

\bibitem[Eysenbach \& Levine(2019)Eysenbach and Levine]{eysenbach2019if}
Eysenbach, B. and Levine, S.
\newblock If maxent rl is the answer, what is the question?
\newblock \emph{arXiv preprint arXiv:1910.01913}, 2019.

\bibitem[Farshidian et~al.(2014)Farshidian, Neunert, and
  Buchli]{farshidian2014learning}
Farshidian, F., Neunert, M., and Buchli, J.
\newblock Learning of closed-loop motion control.
\newblock In \emph{2014 IEEE/RSJ International Conference on Intelligent Robots
  and Systems}, pp.\  1441--1446. IEEE, 2014.

\bibitem[Friston(2005)]{friston2005theory}
Friston, K.
\newblock A theory of cortical responses.
\newblock \emph{Philosophical transactions of the Royal Society B: Biological
  sciences}, 360\penalty0 (1456):\penalty0 815--836, 2005.

\bibitem[Friston et~al.(2015)Friston, Rigoli, Ognibene, Mathys, Fitzgerald, and
  Pezzulo]{friston2015active}
Friston, K., Rigoli, F., Ognibene, D., Mathys, C., Fitzgerald, T., and Pezzulo,
  G.
\newblock Active inference and epistemic value.
\newblock \emph{Cognitive neuroscience}, 6\penalty0 (4):\penalty0 187--214,
  2015.

\bibitem[Gl{\"a}scher et~al.(2010)Gl{\"a}scher, Daw, Dayan, and
  O'Doherty]{glascher2010states}
Gl{\"a}scher, J., Daw, N., Dayan, P., and O'Doherty, J.~P.
\newblock States versus rewards: dissociable neural prediction error signals
  underlying model-based and model-free reinforcement learning.
\newblock \emph{Neuron}, 66\penalty0 (4):\penalty0 585--595, 2010.

\bibitem[Gu et~al.(2016)Gu, Lillicrap, Sutskever, and Levine]{gu2016continuous}
Gu, S., Lillicrap, T., Sutskever, I., and Levine, S.
\newblock Continuous deep q-learning with model-based acceleration.
\newblock In \emph{International Conference on Machine Learning}, pp.\
  2829--2838, 2016.

\bibitem[Haarnoja et~al.(2018)Haarnoja, Zhou, Abbeel, and
  Levine]{haarnoja2018soft}
Haarnoja, T., Zhou, A., Abbeel, P., and Levine, S.
\newblock Soft actor-critic: Off-policy maximum entropy deep reinforcement
  learning with a stochastic actor.
\newblock \emph{arXiv preprint arXiv:1801.01290}, 2018.

\bibitem[Hjelm et~al.(2016)Hjelm, Salakhutdinov, Cho, Jojic, Calhoun, and
  Chung]{hjelm2016iterative}
Hjelm, D., Salakhutdinov, R.~R., Cho, K., Jojic, N., Calhoun, V., and Chung, J.
\newblock Iterative refinement of the approximate posterior for directed belief
  networks.
\newblock In \emph{Advances in Neural Information Processing Systems}, pp.\
  4691--4699, 2016.

\bibitem[Hoffman et~al.(2013)Hoffman, Blei, Wang, and
  Paisley]{hoffman2013stochastic}
Hoffman, M.~D., Blei, D.~M., Wang, C., and Paisley, J.
\newblock Stochastic variational inference.
\newblock \emph{The Journal of Machine Learning Research}, 14\penalty0
  (1):\penalty0 1303--1347, 2013.

\bibitem[Jordan et~al.(1999)Jordan, Ghahramani, Jaakkola, and
  Saul]{jordan1999introduction}
Jordan, M.~I., Ghahramani, Z., Jaakkola, T.~S., and Saul, L.~K.
\newblock An introduction to variational methods for graphical models.
\newblock \emph{Machine learning}, 37\penalty0 (2):\penalty0 183--233, 1999.

\bibitem[Kim et~al.(2018)Kim, Wiseman, Miller, Sontag, and Rush]{kim2018semi}
Kim, Y., Wiseman, S., Miller, A.~C., Sontag, D., and Rush, A.~M.
\newblock Semi-amortized variational autoencoders.
\newblock \emph{arXiv preprint arXiv:1802.02550}, 2018.

\bibitem[Kingma \& Welling(2013)Kingma and Welling]{kingma2013auto}
Kingma, D.~P. and Welling, M.
\newblock Auto-encoding variational bayes.
\newblock \emph{arXiv preprint arXiv:1312.6114}, 2013.

\bibitem[Krishnan et~al.(2017)Krishnan, Liang, and
  Hoffman]{krishnan2017challenges}
Krishnan, R.~G., Liang, D., and Hoffman, M.
\newblock On the challenges of learning with inference networks on sparse,
  high-dimensional data.
\newblock \emph{arXiv preprint arXiv:1710.06085}, 2017.

\bibitem[Kurutach et~al.(2018)Kurutach, Clavera, Duan, Tamar, and
  Abbeel]{kurutach2018model}
Kurutach, T., Clavera, I., Duan, Y., Tamar, A., and Abbeel, P.
\newblock Model-ensemble trust-region policy optimization.
\newblock \emph{arXiv preprint arXiv:1802.10592}, 2018.

\bibitem[Levine(2018)]{levine2018reinforcement}
Levine, S.
\newblock Reinforcement learning and control as probabilistic inference:
  Tutorial and review.
\newblock \emph{arXiv preprint arXiv:1805.00909}, 2018.

\bibitem[Li(2020)]{li2020robot}
Li, S.
\newblock Robot playing kendama with model-based and model-free reinforcement
  learning.
\newblock \emph{arXiv preprint arXiv:2003.06751}, 2020.

\bibitem[Marino(2019)]{marino2019predictive}
Marino, J.
\newblock Predictive coding, variational autoencoders, and biological
  connections.
\newblock 2019.

\bibitem[Marino et~al.(2018{\natexlab{a}})Marino, Cvitkovic, and
  Yue]{marino2018general}
Marino, J., Cvitkovic, M., and Yue, Y.
\newblock A general method for amortizing variational filtering.
\newblock In \emph{Advances in Neural Information Processing Systems}, pp.\
  7857--7868, 2018{\natexlab{a}}.

\bibitem[Marino et~al.(2018{\natexlab{b}})Marino, Yue, and
  Mandt]{marino2018iterative}
Marino, J., Yue, Y., and Mandt, S.
\newblock Iterative amortized inference.
\newblock \emph{arXiv preprint arXiv:1807.09356}, 2018{\natexlab{b}}.

\bibitem[Mnih et~al.(2015)Mnih, Kavukcuoglu, Silver, Rusu, Veness, Bellemare,
  Graves, Riedmiller, Fidjeland, Ostrovski, et~al.]{mnih2015human}
Mnih, V., Kavukcuoglu, K., Silver, D., Rusu, A.~A., Veness, J., Bellemare,
  M.~G., Graves, A., Riedmiller, M., Fidjeland, A.~K., Ostrovski, G., et~al.
\newblock Human-level control through deep reinforcement learning.
\newblock \emph{Nature}, 518\penalty0 (7540):\penalty0 529--533, 2015.

\bibitem[Nachum et~al.(2017)Nachum, Norouzi, Xu, and
  Schuurmans]{nachum2017bridging}
Nachum, O., Norouzi, M., Xu, K., and Schuurmans, D.
\newblock Bridging the gap between value and policy based reinforcement
  learning.
\newblock In \emph{Advances in Neural Information Processing Systems}, pp.\
  2775--2785, 2017.

\bibitem[Nagabandi et~al.(2018)Nagabandi, Kahn, Fearing, and
  Levine]{nagabandi2018neural}
Nagabandi, A., Kahn, G., Fearing, R.~S., and Levine, S.
\newblock Neural network dynamics for model-based deep reinforcement learning
  with model-free fine-tuning.
\newblock In \emph{2018 IEEE International Conference on Robotics and
  Automation (ICRA)}, pp.\  7559--7566. IEEE, 2018.

\bibitem[Niv et~al.(2006)Niv, Joel, and Dayan]{niv2006normative}
Niv, Y., Joel, D., and Dayan, P.
\newblock A normative perspective on motivation.
\newblock \emph{Trends in cognitive sciences}, 10\penalty0 (8):\penalty0
  375--381, 2006.

\bibitem[Okada \& Taniguchi(2018)Okada and Taniguchi]{okada2018acceleration}
Okada, M. and Taniguchi, T.
\newblock Acceleration of gradient-based path integral method for efficient
  optimal and inverse optimal control.
\newblock In \emph{2018 IEEE International Conference on Robotics and
  Automation (ICRA)}, pp.\  3013--3020. IEEE, 2018.

\bibitem[Okada \& Taniguchi(2019)Okada and Taniguchi]{okada2019variational}
Okada, M. and Taniguchi, T.
\newblock Variational inference mpc for bayesian model-based reinforcement
  learning.
\newblock \emph{arXiv preprint arXiv:1907.04202}, 2019.

\bibitem[Pich{\'e} et~al.(2018)Pich{\'e}, Thomas, Ibrahim, Bengio, and
  Pal]{piche2018probabilistic}
Pich{\'e}, A., Thomas, V., Ibrahim, C., Bengio, Y., and Pal, C.
\newblock Probabilistic planning with sequential monte carlo methods.
\newblock 2018.

\bibitem[Rao \& Ballard(1999)Rao and Ballard]{rao1999predictive}
Rao, R.~P. and Ballard, D.~H.
\newblock Predictive coding in the visual cortex: a functional interpretation
  of some extra-classical receptive-field effects.
\newblock \emph{Nature neuroscience}, 2\penalty0 (1):\penalty0 79--87, 1999.

\bibitem[Rawlik et~al.(2013)Rawlik, Toussaint, and
  Vijayakumar]{rawlik2013stochastic}
Rawlik, K., Toussaint, M., and Vijayakumar, S.
\newblock On stochastic optimal control and reinforcement learning by
  approximate inference.
\newblock In \emph{Twenty-Third International Joint Conference on Artificial
  Intelligence}, 2013.

\bibitem[Satorras et~al.(2019)Satorras, Akata, and
  Welling]{satorras2019combining}
Satorras, V.~G., Akata, Z., and Welling, M.
\newblock Combining generative and discriminative models for hybrid inference.
\newblock In \emph{Advances in Neural Information Processing Systems}, pp.\
  13802--13812, 2019.

\bibitem[Shu et~al.(2019)Shu, Bui, Whang, and Ermon]{shu2019training}
Shu, R., Bui, H.~H., Whang, J., and Ermon, S.
\newblock Training variational autoencoders with buffered stochastic
  variational inference.
\newblock \emph{arXiv preprint arXiv:1902.10294}, 2019.

\bibitem[Silver et~al.(2016)Silver, Huang, Maddison, Guez, Sifre, Van
  Den~Driessche, Schrittwieser, Antonoglou, Panneershelvam, Lanctot,
  et~al.]{silver2016mastering}
Silver, D., Huang, A., Maddison, C.~J., Guez, A., Sifre, L., Van Den~Driessche,
  G., Schrittwieser, J., Antonoglou, I., Panneershelvam, V., Lanctot, M.,
  et~al.
\newblock Mastering the game of go with deep neural networks and tree search.
\newblock \emph{nature}, 529\penalty0 (7587):\penalty0 484, 2016.

\bibitem[Silver et~al.(2017)Silver, Schrittwieser, Simonyan, Antonoglou, Huang,
  Guez, Hubert, Baker, Lai, Bolton, et~al.]{silver2017mastering}
Silver, D., Schrittwieser, J., Simonyan, K., Antonoglou, I., Huang, A., Guez,
  A., Hubert, T., Baker, L., Lai, M., Bolton, A., et~al.
\newblock Mastering the game of go without human knowledge.
\newblock \emph{Nature}, 550\penalty0 (7676):\penalty0 354--359, 2017.

\bibitem[Sutton(1990)]{sutton1990integrated}
Sutton, R.~S.
\newblock Integrated architectures for learning, planning, and reacting based
  on approximating dynamic programming.
\newblock In \emph{Machine learning proceedings 1990}, pp.\  216--224.
  Elsevier, 1990.

\bibitem[Sutton(1991)]{sutton1991dyna}
Sutton, R.~S.
\newblock Dyna, an integrated architecture for learning, planning, and
  reacting.
\newblock \emph{ACM Sigart Bulletin}, 2\penalty0 (4):\penalty0 160--163, 1991.

\bibitem[Tschantz et~al.(2019)Tschantz, Baltieri, Seth, Buckley,
  et~al.]{tschantz2019scaling}
Tschantz, A., Baltieri, M., Seth, A., Buckley, C.~L., et~al.
\newblock Scaling active inference.
\newblock \emph{arXiv preprint arXiv:1911.10601}, 2019.

\bibitem[Walsh et~al.(2020)Walsh, McGovern, Clark, and
  O'Connell]{walsh2020evaluating}
Walsh, K.~S., McGovern, D.~P., Clark, A., and O'Connell, R.~G.
\newblock Evaluating the neurophysiological evidence for predictive processing
  as a model of perception.
\newblock \emph{Annals of the New York Academy of Sciences}, 1464\penalty0
  (1):\penalty0 242, 2020.

\bibitem[Wang \& Ba(2019)Wang and Ba]{wang2019exploring}
Wang, T. and Ba, J.
\newblock Exploring model-based planning with policy networks.
\newblock \emph{arXiv preprint arXiv:1906.08649}, 2019.

\bibitem[Weiss \& Freeman(2000)Weiss and Freeman]{weiss2000correctness}
Weiss, Y. and Freeman, W.~T.
\newblock Correctness of belief propagation in gaussian graphical models of
  arbitrary topology.
\newblock In \emph{Advances in neural information processing systems}, pp.\
  673--679, 2000.

\bibitem[Williams et~al.(2017)Williams, Wagener, Goldfain, Drews, Rehg, Boots,
  and Theodorou]{williams2017information}
Williams, G., Wagener, N., Goldfain, B., Drews, P., Rehg, J.~M., Boots, B., and
  Theodorou, E.~A.
\newblock Information theoretic mpc for model-based reinforcement learning.
\newblock In \emph{2017 IEEE International Conference on Robotics and
  Automation (ICRA)}, pp.\  1714--1721. IEEE, 2017.

\bibitem[Winn \& Bishop(2005)Winn and Bishop]{winn2005variational}
Winn, J. and Bishop, C.~M.
\newblock Variational message passing.
\newblock \emph{Journal of Machine Learning Research}, 6\penalty0
  (Apr):\penalty0 661--694, 2005.

\bibitem[Ziebart(2010)]{ziebart2010modeling}
Ziebart, B.~D.
\newblock Modeling purposeful adaptive behavior with the principle of maximum
  causal entropy.
\newblock 2010.

\end{thebibliography}
\bibliographystyle{icml2020}


\appendix
\onecolumn

\section{Full Algorithm}
\label{ap:algo}

\begin{algorithm}[H]
  \label{algo:chi}
  \SetAlgoLined
     \DontPrintSemicolon
     \textbf{Input:} Planning horizon $H$ | Optimisation iterations $I$ | Number of samples $K$ | Current state $\rs_t$ |  Transition distribution $p_{\lambda}(\rs_{t+1}|\rs_t, \ra_t)$ | Amortisation function $f_{\phi}(\cdot)$
     \BlankLine
     \textbf{Amortised Inference}: \\
    $p_{\phi}(\tau) = \delta(\rs_t) \prod_{t'=t}^T p_{\lambda}(\rs_{t'+1}|\rs_{t'}, \ra_{t'}) q_{\phi}(\ra_{t'}|\rs_{t'}; \theta)$ \\
     Extract $\theta^{(1)} = \{\mu_{t:T}, \sigma_{t:T}^2 \}$ from $p_{\phi}(\tau) $ \\
     Initialise $q(\rah; \theta)$ with parameters $\theta^{(1)}$
     \BlankLine
     \textbf{Iterative Inference}: \\
     \For{$\mathrm{optimisation \ iteration} \ i = 1...I$}{
        Sample $K$ action sequences $\{(\rah)_k \sim q(\rah; \theta)\}^K_{k=1}$ \\
        Initialise particle weights $\mathbf{W}^{(i)} := \{w_{k}^{(i)}\}^{K}_{k=1}$ \\
        \For{$\mathrm{action \ sequence} \ k = 1...K$}{
         $ w_k^{(i+1)} \leftarrow \frac{\mathcal{W}\big((\rah)_k\big) \cdot q^{(i)}\big((\rah)_k; \theta \big)}{\sum_{j = 1}^K \Big[\mathcal{W}\big((\rah)_j\big)\ \cdot q^{(i)}\big((\rah)_j; \theta \big)\Big]}$ \\
        $\theta^{(i+1)} \leftarrow \texttt{refit}\big(\mathbf{W}^{(i+1}\big)$
     }
  }
  \BlankLine
  Extract $\mu_{t:T}$ from $q(\rah; \theta)$ \\
  \textbf{return} $\mu_t$
  \caption{Inferring actions via CHI}
\end{algorithm}

\section{Model Details}
\label{ap:model-details}

The proposed CHI algorithm requires a model of the transition dynamics $p_{\lambda}(\rs_{t+1}|\rs_{t}, \ra_t)$. 
This model appears in the iterative inference algorithm, where it is used to evaluate the expected trajectory of states $\rsh$, given some sampled action sequence $(\rah)_k$.
The model also appears in the amortised inference algorithm (Eq. \ref{eq:policy-dist}), where it is again used to calculate the expected trajectory of states under an amortised policy $q_{\phi}(\rat|\rsh; \theta)$.

Rather than treat $\lambda$ as a point estimate, we consider it to a random variable which must be inferred based on the available dataset $\mathcal{D}$.
Here, we utilise an ensemble approach to approximating $p(\lambda|\mathcal{D})$ \cite{chua2018deep, kurutach2018model}.
This approach approximates $p(\lambda|\mathcal{D})$ as a set of particles $p(\lambda|\mathcal{D}) \simeq \frac{1}{E}\sum_{i}^K \delta(\lambda - \lambda_i)$, where $E$ is the number of networks in the ensemble and $\delta$ is the Dirac delta function. 
Each particle $\lambda_i$ is optimised to maximise $\log p(\lambda_i|\mathcal{D}) \propto \log p(\mathcal{D}|\lambda_i)p(\lambda_i)$, and where a uniform prior over $\lambda_i$ is assumed. The model is updated after each episode, iterating over the available dataset for 10 epochs in batches of 50. An ensemble of 5 is used, where each member of the ensemble is a 3 layer neural network with 350 nodes, which is trained to predict a Gaussian distribution over the \emph{change} in state, as opposed to directly predicting the next state. 

\section{Parameters}
\label{ap:parameters}

We utilise a relatively small number of parameters relative to previous work. We consider a planning horizon $H$ of 7, 500 samples for the iterative inference procedure. During training, action noise $\epsilon \sim \mathcal{N}(0, 0.3)$ is added to actions to promote exploration. We use a hidden size of 256 for the value and Q-networks used in SAC. We do not use an adaptive $\alpha$ for SAC and instead set it to a constant value of $0.2$.

\section{Experiment Details}
\label{ap:experimental-details}

\subsection{Didactic Experiment}
For the didactic experiment, we use a simple 2D point mass environment, where the agent must simply navigate to a goal.
The agent can control its $x$ and $y$ velocity, $\ra = (\Delta x, \Delta y)$, with a maximum magnitude of $||\ra|| = 0.05$. 
The environment is a grid of shape $[0,1]^2$ that contains a wall with a small opening, making a direct path to the goal impossible.
The agent starts at $(x = 0, y = 0)$ and the goal is at $\mathbf{g} = (x = 1, y = 1.0)$. The reward function is$r(\rs_t, \ra_t) = 1 - ||\rs_t - \mathbf{g}||^2$ which rewards the agent for navigating towards the goal. 

\subsection{Continious Control Benchmark}
We utilise the Half Cheetah environment ($\mathcal{S} \subseteq \mathbb{R}^17 \mathcal{A} \subseteq \mathbb{R}^6$) which describes a running planar biped. We consider a running task, where a reward of $v - 0.1 ||a||^2$ is received, where $v$ is the agent's velocity and $a$ are the agent's actions. Each episode consists of 100 steps, where an action repeat of 4 is used.  

\section{Evidence Lower Bound Derivation}
\label{ap:bound}

\begin{equation}
  \label{eq:bound-ap}
    \begin{aligned}
      \log  p(\optimalt) & = \log \int p\left(\tau|\optimalt \right) d \rs_{1: T} d \ra_{1: T} \\
      =&  \log \int p\left(\tau|\optimalt \right) \frac{q\left(\tau \right)}{q\left(\tau \right)} d \rs_{1: T} d \ra_{1: T} \\
      =& \log \mathbb{E}_{q\left(\tau \right)} \frac{p\left(\tau|\optimalt \right)}{q\left(\tau \right)} \\
      \geq & \  \mathbb{E}_{q\left(\tau \right)}\big[\log p\left( \tau  | \optimalt  \right) -\log q\left(\tau \right)\big]
      \end{aligned}
  \end{equation}

\section{Evidence Lower Bound Simplification}
\label{ap:simplifcation}

Following the main text, we define $q(\rs_1) = p(\rs_1)$ and $q(\rs_{t+1}| \rs_t, \ra_{t}) = p_{\lambda}(\rs_{t+1}| \rs_t, \ra_{t})$, allowing us to simplify Equation \ref{eq:bound-ap} as follows:

\begin{equation}
\label{eq:simple-bound-ap}
  \begin{aligned}
    & \mathbb{E}_{q(\tau)}\big[\log p\left( \tau | \optimalt  \right)- \log q\left(\tau \right)\big] \\
    &= \mathbb{E}_{q(\tau)}\big[ \log p(\rs_1) + \log p(\optimalt | \tau) + \log p_{\lambda}(\rs_{2:T}|\rst, \rat) \\
    & - \log p(\rs_1) - \log q_{\theta}(\rat|\rst) - \log p_{\lambda}(\rs_{2:T}|\rst, \rat)\big] \\
    &=\mathbb{E}_{q(\tau)}\big[\log p(\optimalt| \tau) - \log q_{\theta}(\rat|\rst)\big]
\end{aligned}
\end{equation}

where the last line is derived from the fact that the terms $p_{\lambda}(\rs_{t+1}| \rs_t, \ra_{t})$ and $p(\rs_1)$ appear on both the numerator and denominator.

\begin{figure}
  \begin{center}
        \includegraphics[width=0.35\textwidth]{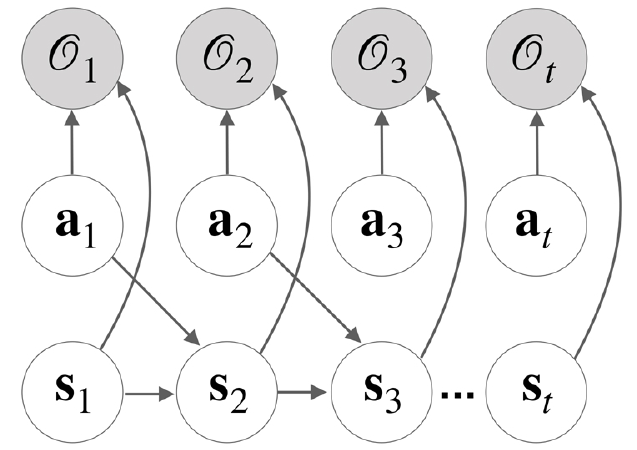}
  \end{center}
  \caption{Graphical model for control as inference.}
\label{fig:graphical-model}
\end{figure}

\section{The Data Bias Issue}
\label{ap:data-bias}
Training the amortised algorithm from data generated by the CHI algorithm poses a significant issue, which we here refer to as the \emph{data bias} issue. Because iterative planning algorithms consider hundreds of potential actions before selecting one to be executed, they have an inherent bias to avoiding sub-optimal actions. The resulting dataset is thus highly biased towards good actions, meaning that information about sub-optimal actions does not get propagated to the amortised algorithm. This poses an issue -- even for off-policy algorithms such as SAC. To overcome this, we train the amortised algorithm on the counterfactual rollouts generated by the iterative planning algorithm. This procedure generates a huge amount of (suboptimal and optimal) data that would otherwise be discarded. Empirically, we found this method to be crucial for successful learning. 

\section{Related Work}
\label{ap:related-work}

\paragraph{Combining model-based and model-free RL} A number of methodologies exist for combining model-free and model-based RL \cite{li2020robot, che2018combining}.
Previous work has considered using a learned model to generate additional data for training a model-free policy \cite{gu2016continuous, sutton1990integrated, sutton1991dyna}.
In \cite{chebotar2017combining}, the authors consider linear-Gaussian controllers as policies and derive both model-based and model-free updates. 
In \cite{farshidian2014learning}, the authors consider a similair initalisation approach to our own, but use a model-based algorithm to initialize a model-free algorithm. 
This is in contrast to our approach, where the model-free policy initializes the model-based planning algorithm. 
The initalization method used in the current paper mirrors the use of policy networks to generate proposals for the Monte-Carlo tree search in AlphaGo \cite{silver2016mastering, silver2017mastering}. 
Several papers look to use the learned model to initialize a model-free policy \cite{nagabandi2018neural}.

\paragraph{Combining Amortised \& Iterative Inference} The idea of combing amortised and iterative inference has been explored previously the context of unsupervised learning. 
Such approaches look to retain the computational efficiency of amortised inference models while incorporating the more powerful capabilities of iterative inference. 
The semi-amortised variational autoencoder was introduced in \cite{kim2018semi}, which also employs amortised inference to initialize a set of variational parameters, which are then refined using iterative inference. 
The authors demonstrate that this approach helps overcome the `posterior collapse' phenomenon, which describes when the latent code of the auto-encoder is ignored and presents a common issue when training variational autoencoders.
An iterative amortised inference algorithm was proposed by \cite{marino2018iterative}, where posterior estimates provided by amortised inference are iteratively refined by repeatedly encoding gradients. 
It was demonstrated that this approach helps overcome the \emph{amortisation gap} \cite{krishnan2017challenges, cremer2018inference}, which describes the tendency for amortised inference models to not reach fully optimised posterior estimates, likely due to the significant restriction of optimising a direct (and generally feed-forward) mapping from data to posterior parameters.  
This iterative amortised inference model was later applied to variational filtering \cite{marino2018general}.
In \cite{satorras2019combining}, the authors propose a hybrid inference scheme for combing generative and discriminative models, which is applied to a Kalman Filter, demonstrating an improved accuracy relative to the constituent inference systems. 
The biological plausibility of hybrid inference schemes has been explored in the context of perception \cite{marino2019predictive}, utilising the predictive coding framework from cognitive neuroscience \cite{rao1999predictive, friston2005theory, walsh2020evaluating}.
A hybrid inference approach which iteratively refines amortised predictions has also been explored in \cite{hjelm2016iterative, krishnan2017challenges, shu2019training}.

\end{document}